\pdfoutput=1

\documentclass[11pt]{article}

\usepackage[]{acl}
\usepackage{graphicx}
\usepackage{times}
\usepackage{xcolor}
\usepackage{latexsym}
\usepackage{hyperref}
\usepackage{lipsum}
\usepackage{multirow}
\usepackage{breqn}
\usepackage[T1]{fontenc}
\usepackage{algpseudocode}
\usepackage{algorithm}
\usepackage{caption}
\usepackage{subcaption}
\usepackage{booktabs}
\usepackage{pgfplots}
\pgfplotsset{compat=1.18}
\usepackage{wasysym}
\usepackage{tabularx}

\newcommand\blfootnote[1]{%
  \begingroup
  \renewcommand\thefootnote{}\footnote{#1}%
  \addtocounter{footnote}{-1}%
  \endgroup
}

\usepackage{fontawesome5}

\usepackage[utf8]{inputenc}

\usepackage{microtype}

\title{Privacy-Preserving Social Mental Health Text Classification for Mitigating Online Abuse and Harms}
\title{Assessing Social Mental Health Risks in a Privacy-Preserving Manner}
\title{Privacy Aware Question-Answering System for Online Mental Health Risk Assessment}

\author{Prateek Chhikara\thanks{~~  Corresponding author}~~\textsuperscript{\dag},~~Ujjwal Pasupulety\textsuperscript{\dag},~~John Marshall, \\\textbf{Dhiraj Chaurasia }and \textbf{Shweta Kumari} \\
        University of Southern California, Los Angeles, USA \\
        \texttt{\{pchhikar,upasupul,jjmarsha,dchauras,shwetaku\}@usc.edu}
}
\begin{document}
\maketitle

\begin{abstract}
Social media platforms have enabled individuals suffering from mental illnesses to share their lived experiences and find the online support necessary to cope. However, many users fail to receive genuine clinical support, thus exacerbating their symptoms. Screening users based on what they post online can aid providers in administering targeted healthcare and minimize false positives. Pre-trained Language Models (LMs) can assess users' social media data and classify them in terms of their mental health risk. We propose a Question-Answering (QA) approach to assess mental health risk using the Unified-QA model on two large mental health datasets. To protect user data, we extend Unified-QA by anonymizing the model training process using differential privacy. Our results demonstrate the effectiveness of modeling risk assessment as a QA task, specifically for mental health use cases. Furthermore, the model's performance decreases by less than 1\% with the inclusion of differential privacy. The proposed system's performance is indicative of a promising research direction that will lead to the development of privacy-aware diagnostic systems.
\blfootnote{\dag ~~These authors contributed equally to this work}    
\end{abstract}

\section{Introduction}
In recent years, Natural Language Processing (NLP) has emerged as a powerful field of study that focuses on the interaction between human language and computational systems \cite{singh2020ensemble}. Mental health is a crucial aspect of overall well-being, and gaining insights into individuals' mental states has become an increasingly important area of study. NLP techniques have been useful in identifying text markers that indicate an individual's mental well-being \cite{mlrev2}. 
Social media websites, such as Twitter and Reddit, provide a wealth of textual data that offers a unique opportunity to analyze the mental health status of their users at scale, enabling the exploration of patterns, trends, and potential interventions \cite{Skaik2020UsingSM}.
Assessing users' mental health risk can be reduced to a basic text classification task, where the Transformer architecture \cite{attn} has demonstrated state-of-the-art performance. BERT \cite{Devlin2019BERTPO} encodings have been utilized for training a variety of mental health risk detection systems \cite{jiang-etal-2020-detection,Nisa2021TowardsTL,Zeberga2022}. BERT models fine-tuned on social media data \cite{ji-etal-2022-mentalbert,Murarka2020ClassificationOM} are able to classify at-risk individuals with high accuracy.

\begin{figure*}[!htbp]
\centering
\begin{subfigure}{.42\textwidth}
  \centering
  \includegraphics[width=\linewidth]{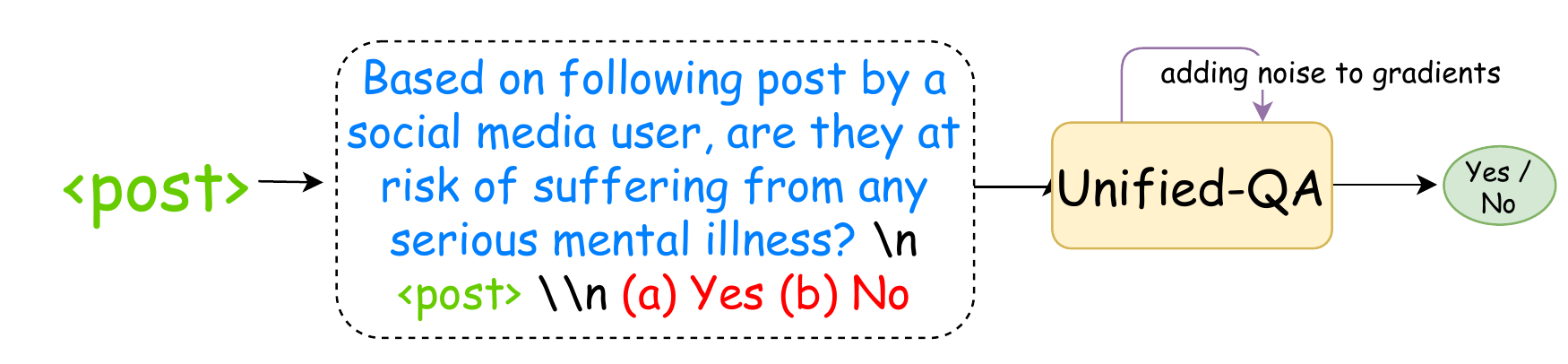}
  \caption{Binary Classification}
  \label{fig:sub1}
\end{subfigure}%
\begin{subfigure}{.55\textwidth}
  \centering
  \includegraphics[width=\linewidth]{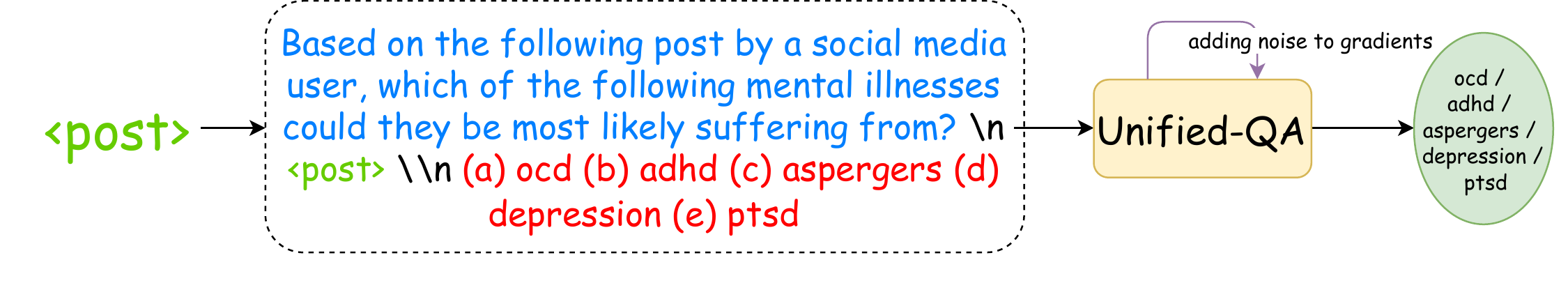}
  \caption{Multi Classification}
  \label{fig:sub2}
\end{subfigure}
\caption{Proposed Pipeline.}
\label{proposed_work}
\end{figure*}
However, advances in text classification models have stagnated with the advent of BERT encodings. Posing the risk assessment problem as a QA task is more analogous to consulting a trained clinician \cite{medqa}. QA systems built using BERT have been used for public education on topics in mental health \cite{mhqa}. 
 Nearly 30\% of QA healthcare systems focus on mental health applications such as workplace empowerment, screening, effecting behavior change, and reducing smoking/alcohol dependence \cite{qarev}. 
 Multiple-choice QA models demonstrate a promising alternative approach to depression severity estimation even with low amounts of training data \cite{engproc2021007023}. Further development of QA models could lead to better autonomous diagnostic systems. This work proposes the use of AllenAI's Unified-QA model \cite{khashabi-etal-2020-unifiedqa} to assess the mental health risk of users from their social media posts. The research objective is to explore whether QA transformer models are better than text classification transformers at assessing the risk to mental health and modeling language markers that are indicative of specific mental illnesses. We compare Unified-QA to state-of-the-art pre-trained language models that perform text classification on the same data.

Training models on sensitive user data in their raw form makes them non-compliant with data privacy rules which can have serious legal ramifications in the case of unexpected data breaches \cite{dpethics}. By using Differential Privacy, language models can be trained such that they do not memorize the training data, leading to data security and better model generalization \cite{Basu2021BenchmarkingDP,Behnia_2022}. This work also studies the impact of differential private training on QA model performance. The contributions of the paper are as follows.
\begin{enumerate}
\itemsep0em
    \item We approached the text classification task for mental health posts using a QA framework. Specifically, we transformed our input data according to the Unified-QA model and subsequently fine-tuned and evaluated it against one binary dataset and one multi-class dataset.
    \item We add noise to the model gradients to prevent the QA model from memorizing the input text, thereby making it differentially private.
    \item Our model's performance is compared against various text classification models, including both traditional classifiers and pre-trained language models with varying degrees of scale, spanning from low to high-parameter models.
\end{enumerate}

\section{Proposed Methodology}
The proposed work consists of fine-tuning the Unified-QA model on two large mental health datasets in a differentially private manner.

\subsection{Mental Health Assessment as a QA task}

Treating a text classification task as a seq2seq QA task involves converting input text-label pairs into a Question-Answer format. This approach can be beneficial for tasks such as sentiment analysis, topic classification, and other text classification tasks as it provides a more structured and interpretable output \cite{qasysrev}. Additionally, it allows for using pre-trained QA transformer models, which can improve performance without requiring large amounts of training data \cite{HAN2021225}. In the proposed work, we have utilized Unified-QA, a single pre-trained QA model that performs exceptionally well across 17 diverse QA datasets with different formats. It demonstrates strong generalization even on unseen datasets, outperforming specialized models trained on individual datasets. Fine-tuning the Unified QA model leads to state-of-the-art performance on six datasets, making it a robust foundation for developing QA systems.

The Unified-QA model is prompted as shown in Figure \ref{proposed_work}, where each post-label pair from the training set is pre-processed by creating a task-specific question prompt to which the post is appended. For outputs, each training sample contains answer options whose values are equal to the set of class labels for each task. This approach enables a more nuanced analysis of mental health-related content, providing a clear rationale for the predicted mental health status. The input and output formatting is consistent for both training and testing.

\subsection{Incorporating Differential Privacy}

Mental health datasets are highly sensitive, containing personal and often confidential information about individuals' private lives. Na\"{\i}vely training language models on such data poses a high risk of input-memorization, allowing adversaries to reconstruct samples using the models and their weights \cite{s1,Carlini2020ExtractingTD}. This work enhances the training process of the Unified-QA model by incorporating differential privacy \cite{dppaper}. By adding controlled noise to the gradients during training \cite{noisygrad, chhikara2023adaptive}, a balance between model accuracy and privacy preservation is achieved. Consequently, individuals' data remains secure within the model, even if an adversary gains access to the model's weights. The proposed approach makes a powerful general-purpose QA system suitable for use in applications that involve sensitive data.

\begin{table*}[!t]
\centering
\caption{Samples from each dataset with their corresponding label.}
\label{dataset_desc}
\begin{tabularx}{\textwidth}{X|X}
\toprule
\textbf{\footnotesize RedditMH} & \textbf{\footnotesize SMK} \\
\midrule
\scriptsize \textbf{\texttt{Positive:}} Fed up with my crazy mother. First she was just angry all the time, now she's angry and mad(schizophrenic) all the time, it's frustrating to not get a proper father in the house and add to that not even a stable mother, unstable is not even half the things she is. My god. She thinks my father is cooking up dark magic portions all the time. It's infuriating to have family like this, especially when half the things in my life doesn't seem to work properly or s**t always happens with me. honestly fed up with always trying to improve my life, exercising, meditating, everything under the sun, and they definitely have effect but it's not satisfactory and I always wonder what'd happen if I just didn't have such a mother, and actually have a father in the house. \newline \newline
\textbf{\texttt{Negative:}} Is it possible to lose fat without losing weight? I used to workout all the time and was pretty muscular (80kg) when I was 17. I quit working out like about 3 months ago because I was working 14h everyday to sustain my s**t and had problems back then. I'm now 75ish kg and I'm slim but have high body fat. I found a pretty good job that is pretty good in terms of hours and finally have time to working out. I'll start rucking with 20kg some 5 miles starting Monday. I have gotten a pretty fat face, and some love handles. My question is, can I lose fat but still maintaining my weight?

&
\scriptsize  \textbf{\texttt{ADHD:}} Please, please help me. The past 2-3 months I have been doing very well managing my symptoms and I have been overall, happy. About 2 weeks ago I noticed myself becoming angry just by my fiancé’s presence… it makes me feel crazy as he is literally my favorite person. I have been SO irrationally angry with him about nothing. I can reason with myself, but can’t help my anger. Not currently on any medications as none I have tried have worked. Been off for 3ish months. I’m also upset because Christmas is my FAVORITE and being p***y is ruining it for me. Please help me  \includegraphics[width=2ex]{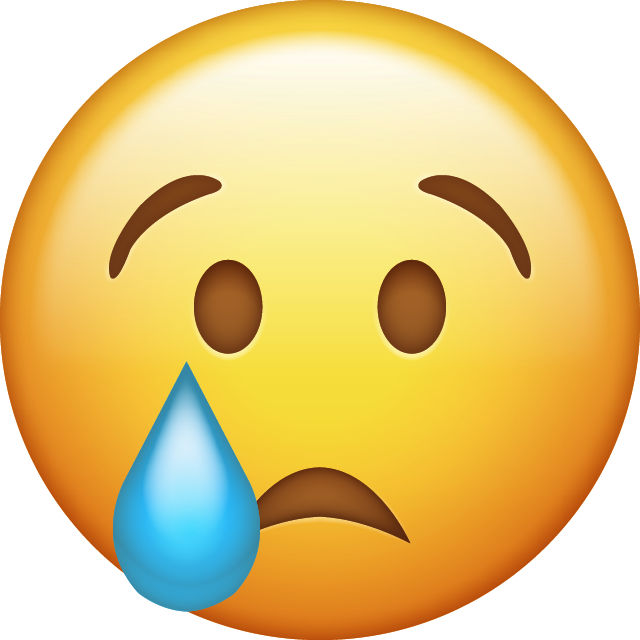} \includegraphics[width=2ex]{crying-emoji.png} \includegraphics[width=2ex]{crying-emoji.png} \includegraphics[width=2ex]{crying-emoji.png} \includegraphics[width=2ex]{crying-emoji.png} \newline \newline
\textbf{\texttt{Depression:}} Does anyone else still manage fully function society's eyes despite severely depressed? I fulltime job I'm never late for, I act happy people believe it, I go gym regularly, I try best maintain friendships even though they're seeming one-sided. To society, I'm fine. What see crying eyes whenever I drive own, I constantly think suicide self harm, I feel like I'm wasting life. When I get home work, I'll sit phone wait go sleep. I interest passions I feel trying push something I used love drains me. But society, I'm fine. \\

\bottomrule
\end{tabularx}
\end{table*}


\subsection{Datasets}
\begin{figure}[!t]
    \centering
\includegraphics[width=\linewidth, trim=.2cm 0.6cm .9cm 0.8cm, clip]{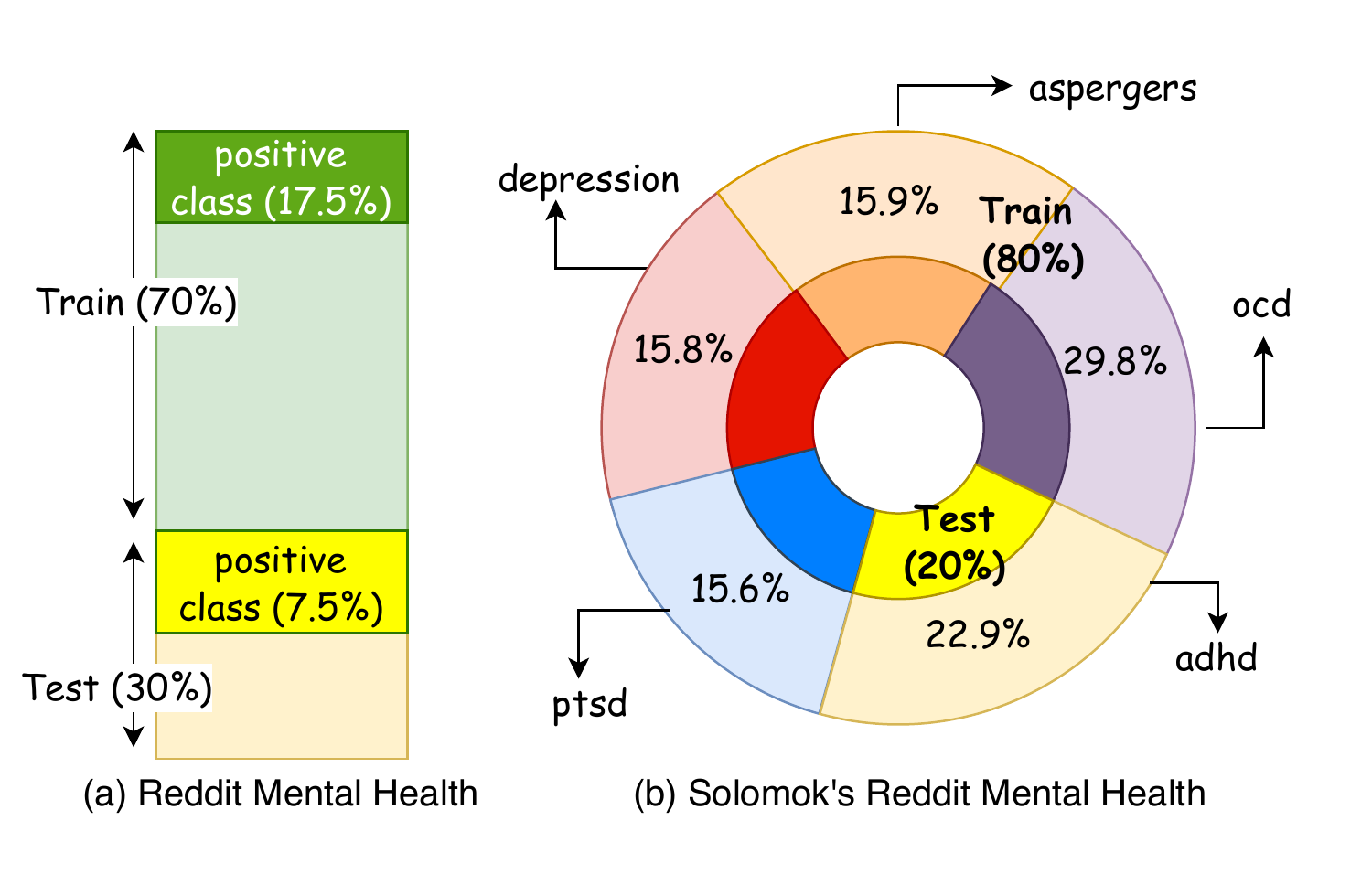}
    \caption{Dataset details.}
    \label{dataset-fig}
\end{figure}

The following public domain datasets were used to fine-tune and evaluate the language models listed in Section \ref{models}. As the datasets are openly available for research purposes, no additional user consent was required to procure them. Table \ref{dataset_desc} shows a snapshot from each dataset and Figure \ref{dataset-fig} shows the train-test splits and class distributions.  \\

\noindent \textbf{Reddit Mental Health Dataset (RedditMH):}
This dataset \cite{rmhdata} contains posts from 28 subreddits collected from 2018-2020, including 15 specific mental health support groups, two broad mental health subreddits, and 11 non-mental health subreddits. For the purpose of this study, instances from broad mental-health (45K) and non-mental health (137K) subreddits are chosen. This was done since LMs perform better at answering clinical questions when pre-trained on open-domain corpora \cite{soni-roberts-2020-evaluation}. \\

\noindent \textbf{Solomonk's Reddit Mental Health (SMK):}
This dataset$\footnote{\url{https://huggingface.co/datasets/solomonk/reddit\_mental\_health\_posts}}$ is comprised of 151K Reddit posts exclusively from five mental health-related subreddits. After cleaning, approximately 88k (depression: 13.8K, OCD: 26.3K, Aspergers: 14.1K, PTSD: 13.7K, and ADHD: 20.3K) entries were considered for experimentation.

\subsection{Models}
The Unified-QA model is fine-tuned and evaluated using the aforementioned datasets. To assess its performance, we compare its results against the following model classes. \\

\noindent \textbf{Baseline classifiers:} The following traditional algorithms were used to define baselines which the LMs improve upon: Logistic Regression (LR), Support Vector Machine (SVM), Decision Tree (DT), Random Forest (RF), AdaBoost, Stochastic Gradient Descent (SGD), Multilayer Perceptron (MLP), ExtRA trees (xTrees) \& Multinomial Naive Bayes (MNb). \\

\noindent \textbf{Low-parameter LMs:} The following transformers (10-25M parameters) were evaluated: (1) \textbf{\textit{roberta-small}}$\footnote{\url{https://huggingface.co/smallbenchnlp/roberta-small}}$ , a distilled version of RoBERTa that can be trained on a single GPU. (2) \citet{nlpie} developed a suite of lightweight transformers specifically for clinical NLP applications. Of these , \textbf{\textit{tiny-clinicalbert}}, \textbf{\textit{clinical-mobilebert}}, \textbf{\textit{bio-mobilebert}}  and \textbf{\textit{tiny-biobert}} \cite{s4} are evaluated. \\

\noindent \textbf{High-parameter LMs:} The following large transformers (>100M parameters) were evaluated on the datasets : (1) \textbf{\textit{PsychBERT}} \cite{psychbert}, a model pre-trained on clinical and biomedical literature. (2) \textbf{\textit{MentalBERT}} and \textbf{\textit{MentalRoBERTa}}, state-of-the-art models which are pre-trained on data from mental health-related subreddits.

\label{models}
\section{Experimental Setup}
\subsection{Model Training}

The baseline classifiers were implemented by training vectorized inputs (TF-IDF, Count, Hash) using scikit-learn on consumer-grade PCs without any hyperparameter tuning or GPU acceleration. The language models were fine-tuned as text-classifiers using Nvidia A10G GPUs with 24 GB GDDR6 memory. They were fine-tuned for five epochs with varying batch sizes using the Hugging Face API with default hyperparameters. Since these models are pre-trained on relevant clinical or social media data, fine-tuning on the evaluation datasets for too many epochs hurts model performance.

The proposed QA models were trained for 20 epochs. The small and base models were trained with 128 and 64 batch sizes, respectively. The input text token length is set to 200. The \textit{Adam} optimizer (lr=$10^{-3}$) with weight decay and a linear learning rate scheduler are used.

\subsection{Differential Privacy}
Fine-tuning a small portion of layers on a subset of the training data can enhance pre-trained model performance on specific tasks and reduce the risk of overfitting. Fine-tuning saves time and resources compared to training a model from scratch on the same dataset, as pre-trained models have already learned multiple language features that are pertinent. Noise, denoted by $noise_{std}$ is sampled from the standard normal distribution ($\mu = 0 , \sigma = 1$) and added during the training phase. The encoder and decoder of the Unified-QA model are frozen while the remaining layers are fine-tuned with 10\% of the evaluation datasets. Clipping the gradients if their norm exceeds a specified maximum value prevents them from exploding during training, which mitigates numerical instability and improves convergence time \cite{Zhang2020Why}. 

\subsection{Evaluating Model Safety}
In order to assess whether the proposed model effectively implements differential privacy, $\epsilon$ and $\delta$ values are computed. $\epsilon$ measures the strength of the privacy guarantee, and $\delta$ is a parameter that bounds the probability of failing to provide the privacy guarantee \cite{s2}. \textit{S} is the sensitivity of the loss function with respect to individual training examples. The $clip_{norm}$ parameter is the maximum L2-norm of the gradient before adding noise. The standard deviation of the maximum noise is computed using Equation \ref{eq1}. \\

\begin{dmath}
\label{eq1}
  maxNoise_{std} \gets clip_{norm} * \sqrt{\frac{2 * total_{epsilon}}{n}} \\ + S * \sqrt{\frac{2 * \log{(\frac{1.25}{\delta})}} {\epsilon}} 
\end{dmath}

where $total_{epsilon}$ is $2\epsilon$ and $n$ represents the number of instances used in the computation. If the standard deviation of the Gaussian noise added ($noise_{std}$) to gradients during training process is less than $maxNoise_{std}$, then the model is differential private.

\section{Results and Analysis}

\subsection{Unified-QA model}
Table \ref{redditmhres} shows the metrics calculated with respect to the positive class (label-1/answer-yes) for the RedditMH (binary choice) dataset. Table \ref{smkres} shows the weighted average results for the SMK (multiple choice) dataset. The Unified-QA model clearly outperforms the top three baselines and low parameter language models. It also performs competitively against state-of-the-art language models, being only slightly outperformed by PsychBERT. The Appendix contains the complete evaluation results for traditional classifiers.

One possible explanation for the superior performance of the QA approach is that it is better suited to the mental health-related text classification task, which requires a deeper understanding of the context and nuances of the text. By transforming the input text into a question, Unified-QA can better capture the semantic relationships and dependencies between the words in the text. Its ability to handle long and complex sentences lends more accurate mental health risk assessments.

\begin{table}[!t]
\caption{Comparison of models on RedditMH dataset.}
\centering
\resizebox{\linewidth}{!}{
\begin{tabular}{ lcccc } 
\toprule
 \textbf{Model} & \textbf{Precision} & \textbf{Recall} & \textbf{F1} \\
\midrule
MLP (TF-IDF) & 87.939 & 88.831 & 88.382\\ 
MLP (Count) & 87.541 & 87.412 & 87.476 \\ 
LR (TF-IDF) & 90.309 & 82.294 & 86.115 \\ 
\midrule
roberta-small & 93.930 & 94.000 & 93.965 \\ 
tiny-clinicalbert & 94.521 & 93.742 & 94.130 \\ 
tiny-biobert & 94.624 & 94.095 & 94.359 \\ 
clinical-mobilebert & 94.036 & 92.522 & 93.273 \\ 
bio-mobilebert & 93.745  & 92.580 & 93.159 \\ 
\midrule
PsychBERT & \textbf{95.911} & 95.389 & 95.649 \\ 
MentalBERT & 94.604 & 92.191 & 93.382 \\ 
MentalRoBERTa & 95.779 & \textbf{95.772} & \textbf{95.775} \\ 
\midrule
Unified-QA-small & 94.947 & \textit{95.564} & \textit{95.258} \\ 
Unified-QA-base & \textit{95.431} & 94.765 & 95.097\\ 
\bottomrule

\end{tabular}}
\label{redditmhres}
\end{table}

\subsection{Differential Private model}
Training the Unified-QA model using differential privacy makes it an ideal system for mental health applications, given the sensitive nature of the training data. The final model's privacy is validated using Equation \ref{eq1}. The F1 score of the Unified-QA-small differential private model dropped by \textbf{0.47\%} and \textbf{0.82\%} (absolute values) on the RedditMH and SMK dataset, respectively. The additional complexity of a multi-class problem challenges the model to make a privacy-performance trade-off even further, which could explain the higher drop in F1. These findings suggest that while differential privacy techniques can help protect user privacy during model training, their impact on model performance is not insignificant. The promise of safeguarding user data while being able to triage at-risk users accurately offers a promising direction of research to reduce the gap in performance.

\begin{table}[!t]
\caption{Comparison of models on SMK dataset.}
\centering
\resizebox{\linewidth}{!}{
\begin{tabular}{ lcccc } 
\toprule
 \textbf{Model} & \textbf{Precision} & \textbf{Recall} & \textbf{F1} \\
\midrule
LR (TF-IDF) & 82.564 & 87.526 & 84.973 \\ 
SGD (TF-IDF) & 82.665 & 87.147 & 84.847\\ 
SGD (Count) & 82.943 & 86.654 & 84.758\\ 
\midrule
roberta-small & 87.520 & 87.590 & 87.500\\ 
tiny-clinicalbert & 88.770 & 88.790 & 88.740 \\ 
tiny-biobert & 88.076 & 87.834 & 87.917\\
clinical-mobilebert & 86.965 & 87.062 & 86.935\\
bio-mobilebert & 86.963 & 87.045 & 86.954\\
\midrule
PsychBERT & \textbf{90.45} & \textbf{90.47} & \textbf{90.42} \\ 
MentalBERT & 88.76 & 88.82 & 88.75\\ 
MentalRoBERTa & 87.29 & 87.39 & 87.32\\
\midrule
Unified-QA-small & 89.455 & 89.525 & 89.419 \\ 
Unified-QA-base & \textit{89.605} & \textit{89.654}  & \textit{89.533} \\ 
\bottomrule

\end{tabular}}
\label{smkres}
\end{table}

\section{Conclusion \& Future Work}
This work aimed to explore the effectiveness of Question-Answering models for mental health risk assessment. The Unified-QA model performed $\sim$2\% better than low-parameter pre-trained language models and produced competitive results when compared to larger state-of-the-art language models. With differential privacy, the model only suffers a sub-one percent drop in performance. This study contributes to the growing body of research on mental health analysis using social media data. It highlights the potential of artificial intelligence techniques to improve our understanding of mental health and support the development of effective interventions.

Some limitations in the design of this study offer directions for further work to improve performance. Since the RedditMH dataset uses posts from general mental health subreddits, the binary-choice QA model was not exposed to finer-grained data from illness-specific subreddits, which could lead to better assessments. Also, the multiple-choice QA model forces an answer from the list of five diseases even if there are no distinct language markers of any disease. Future work will combine binary and multi-choice QA models to filter at-risk individuals. Two-stage QA models will serve as effective screening tools to ultimately provide care to individuals who need it the most.

\section*{Acknowledgements}
We want to express our gratitude to \textit{beam.cloud}\footnote{\url{https://www.beam.cloud}}, an ML Infrastructure startup for generously providing the necessary compute resources to train and evaluate our models. We greatly appreciate their commitment to advancing the field of natural language processing and their willingness to support our research endeavors.

\bibliography{custom}
\bibliographystyle{acl_natbib}

\clearpage
\onecolumn
\section*{Appendix}
\noindent \textbf{Comparison on Traditional ML Classifiers}\\

\begin{table*}[!h]
  \centering
  \caption{Reddit MH: Precision, recall, and F1-score for 9 models on 3 vectorization techniques. x - did not converge within a reasonable time period}
  \label{tab:results}
    \resizebox{\textwidth}{!}{
  \begin{tabular}{lccc|ccc|ccc}
    \toprule
    \multirow{1}{*}{Model} & \multicolumn{3}{c|}{Count Vectorizer} & \multicolumn{3}{c|}{TF-IDF Vectorizer} & \multicolumn{3}{c}{Hashing Vectorizer} \\
    \cmidrule(lr){2-4} \cmidrule(lr){5-7} \cmidrule(lr){8-10}
     & Precision & Recall & F1-score & Precision & Recall & F1-score & Precision & Recall & F1-score \\
    \midrule
    Logistic Regression & 91.05 & 78.993 & 84.594 & 90.309 & 82.294 & \textbf{86.115} & 87.337 & 78.147  & 82.487 \\
    Decision Tree & 74.179 & 75.897 & 75.027 & 74.95 & 76.471 & 75.702 & 69.265 & 71.021 & 70.133 \\
    Random Forest & 91.875 & 77.662 & 84.173 & 91.229 & 78.853 & 84.591 & 92.841 & 69.896 & 79.752 \\
    Ada Boost & 88.097 & 69.662 & 77.802 & 88.179 & 72.235 & 79.415 & 83.667 & 69.006 & 75.634  \\
    Gradient Boosting & 90.417 & 68.618 & 78.023 & 90.436 & 70.919 & 79.497 & 88.841 & 67.088 & 76.447 \\
    Extra Trees & 92.768 & 74.985 & 82.934 & 93.417 & 74.713 & 83.025 & 96.423 & 58.087 & 72.5 \\
    Multinomial NB & 72.013 & 86.765 & 78.703 & 93.003 & 64.22 & 75.978 & x & x & x \\
    SGD Classifier & 91.476 & 79.463 & 85.048 & 90.592 & 80.506 & 85.253 & 88.352 & 74.684 & 80.945 \\
    Multi-Layer Perceptron & 87.541 & 87.412 & \textbf{87.476} & 87.939 & 88.831 & \textbf{88.382} & x & x & x \\
    \bottomrule
  \end{tabular}}
\end{table*}

\begin{table*}[!h]
  \centering
  \caption{SMK: Precision, recall, and F1-score for 9 models on 3 vectorization techniques. x - did not converge within a reasonable time period}
  \label{tab:results}
    \resizebox{\textwidth}{!}{
  \begin{tabular}{lccc|ccc|ccc}
    \toprule
    \multirow{1}{*}{Model} & \multicolumn{3}{c|}{TF-IDF Vectorizer} & \multicolumn{3}{c|}{Count Vectorizer} & \multicolumn{3}{c}{Hashing Vectorizer} \\
    \cmidrule(lr){2-4} \cmidrule(lr){5-7} \cmidrule(lr){8-10}
     & Precision & Recall & F1-score & Precision & Recall & F1-score & Precision & Recall & F1-score \\
    \midrule
    Logistic Regression & 82.564 & 87.526 & \textbf{84.973} & 80.325 & 89.005 & 84.442 & 82.178 & 85.403 & 83.759\\
    SVC & x & x & x & 75.667 & 90.313 & 82.344 & x & x & x\\
    Decision Tree & 74.832 & 78.237 & 76.497 & 73.74 & 77.934 & 75.779 & 72.061 & 75.64 & 73.807 \\
    Random Forest & 76.999 & 88.91 & 82.527 & 77.838 & 87.223 & 82.264 & 74.641 & 89.668 & 81.467\\
    Ada Boost & 82.766 & 62.275 & 71.073 & 80.932 & 67.829 & 73.804 & 77.396 & 69.194 & 73.066 \\
    Gradient Boosting & 83.677 & 81.63 & 82.64 & 79.119 & 86.123 & 82.473 & 79.252 & 85.953 & 82.466 \\
    Extra Trees & 73.5 & 91.014 & 81.325 & 73.688 & 90.256 & 81.135  & 66.807 & 93.213 & 77.831  \\
    SGD Classifier & 82.665 & 87.147 & \textbf{84.847} & 82.943 & 86.654 & \textbf{84.758} & 82.994 & 84.284 & 83.634\\
    Multi-Layer Perceptron & 81.376 & 86.806 & 84.003 & 79.078 & 86.483 & 82.615 & 80.964 & 84.095 & 82.5\\
    \bottomrule
  \end{tabular}}
\end{table*}

\end{document}